\documentclass[10pt,twocolumn,letterpaper]{article}

\usepackage[pagenumbers]{cvpr} 

\usepackage{multirow, nicefrac}
\usepackage{amssymb}
\usepackage{makecell}
\usepackage{xcolor}
\definecolor{lightblue}{RGB}{88,94,170}
\usepackage{graphicx}
\usepackage{array}
\usepackage{amsmath}

\definecolor{cvprblue}{rgb}{0.21,0.49,0.74}
\usepackage[pagebackref,breaklinks,colorlinks,allcolors=cvprblue]{hyperref}

\title{FSBench: A Figure Skating Benchmark for Advancing Artistic Sports Understanding}

\author{Rong Gao$^1$, Xin Liu$^{1}$\thanks{Corresponding author. }, Zhuozhao Hu$^2$, Bohao Xing$^1$, Baiqiang Xia$^3$, Zitong Yu$^{4}$, Heikki Kälviäinen$^{1,5,6}$\\
$^1$Lappeenranta-Lahti University of Technology LUT, Finland\\
$^2$Tianjin University, China
$^3$AMD Silo AI, Finland
$^4$Great Bay University, China\\
$^5$Rensselaer Polytechnic Institute, USA
$^6$Brno University of Technology, Czech Republic \\
\tt\small \{rong.gao, xin.liu, bohao.xing, heikki.kalviainen\}@lut.fi\\ \tt\small huzhuozhao@tju.edu.cn, baiqiang.xia@amd.com, zitong.yu@ieee.org}

\begin{document}
\maketitle
\begin{figure*}[t]
\begin{center}
\includegraphics[width=14cm]{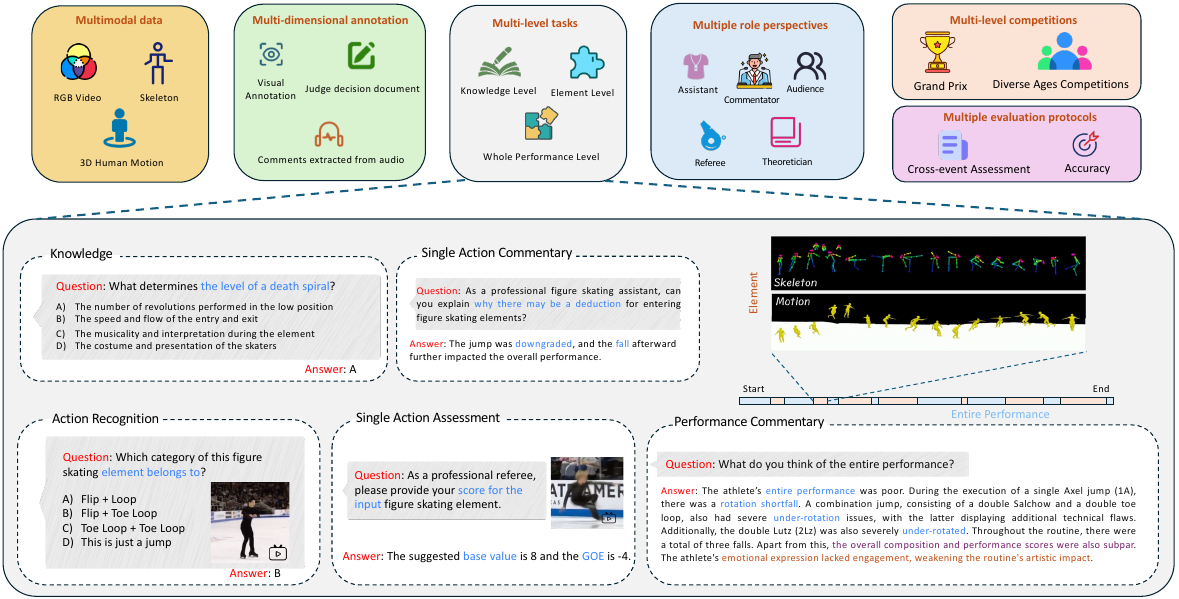}
\end{center}
\vspace{-1em}
\caption{To systematically assess whether models understand figure skating which is an artistic sport, FSBench sources its data from multi-level competitions, incorporating multimodal data, multidimensional annotations, and diverse evaluation protocols. The tasks are structured with multiple roles and levels, ranging from prior knowledge testing, single action recognition, single action assessment, and commentary to the evaluation of entire performances. Additionally, to better evaluate the artistic dimension, our annotations include artistry-related scores, along with corresponding descriptions in the ground truth for both single action assessment and overall performance evaluation tasks. Detailed artistic expression information, such as emotional expression, is usually included in commentary annotations, as shown in the bottom-right corner.}
\vspace{-1em}
\label{fig:fsintro}
\end{figure*}

\begin{abstract}
Figure skating, known as the ``\textbf{Art on Ice},'' is among the most artistic sports, challenging to understand due to its blend of technical elements (like jumps and spins) and overall artistic expression. Existing figure skating datasets mainly focus on single tasks, such as action recognition or scoring, lacking comprehensive annotations for both technical and artistic evaluation. Current sports research is largely centered on ball games, with limited relevance to artistic sports like figure skating. To address this, we introduce \textbf{FSAnno}, a large-scale dataset advancing artistic sports understanding through figure skating. FSAnno includes an open-access training and test dataset, alongside a benchmark dataset, \textbf{FSBench}, for fair model evaluation. FSBench consists of \textbf{FSBench-Text}, with multiple-choice questions and explanations, and \textbf{FSBench-Motion}, containing multimodal data and Question and Answer (QA) pairs, supporting tasks from technical analysis to performance commentary. Initial tests on FSBench reveal significant limitations in existing models’ understanding of artistic sports. We hope FSBench will become a key tool for evaluating and enhancing model comprehension of figure skating.
\end{abstract}    
\section{Introduction}
\label{sec:intro}

Understanding figure skating presents unique challenges due to its intricate fusion of athletic prowess and artistic expression~\cite{hanley2000perennial}. 
Unlike traditional competitive sports, the evaluation of artistic sports is based \textit{not only on technical skill but also on emotional expression and the fluidity of performance}. This subjectivity makes understanding and assessing figure skating more complex.

Current figure skating research in computer vision can be broadly categorized into three areas: i) \textbf{Action Recognition}~\cite{liu2020fsd}, which classifies basic elements such as jumps and spins but fails to capture how these elements contribute to the overall performance; ii) \textbf{Temporal Action Segmentation}\cite{liu2021temporal}, which divides a complete routine into multiple elements without providing individual or holistic evaluations; and iii) \textbf{Action Quality Assessment/Scoring}~\cite{pirsiavash2014assessing, xu2019learning, xia2023skating, du2023learning, liu2023fine}, which predicts the final score for single elements or full performances without integrating them for a comprehensive assessment. 
Despite progress in these areas, existing datasets are insufficient for human-like understanding of figure skating. They focus on isolated tasks and lack the ability to connect individual elements with the overall performance quality, as shown in Table~\ref{tab:dataset}. More importantly, previous studies and datasets on figure skating, similar to those focused on generic or conventional sports movements, primarily emphasize action recognition and score evaluation, overlooking the unique artistry and expressiveness of figure skating. \textit{They ignore artistic elements such as rhythm, fluidity, and emotional expression, which are crucial to the overall performance quality in figure skating.}
This will limit current mainstream MLLM approaches from performing multi-task understanding in figure skating.
Ideally, understanding figure skating requires first evaluating the quality of each element and then analyzing how these elements collectively enhance or detract from the entire performance~\cite{hanley2000perennial, xu2019learning, xia2023skating}. 
To address these limitations and better embrace comprehensive, multifaceted, and in-depth approaches to the training and evaluation of figure skating understanding, we propose \textbf{FSAnno}, a comprehensive dataset with multi-level annotations covering both technical elements and overall performance. FSAnno provides two types of artistry annotations: scoring and textual commentary. In addition to providing data for training and validation, we also offer a separate multi-level evaluation benchmark, \textbf{FSBench}, which provides a structured approach to assess figure skating with human-like depth and insight as shown in Figure~\ref{fig:fsintro}, including assessing rhythm, emotional expression, and movement coherence over extended time spans.

As sports understanding ~\cite{shih2017survey, cust2019machine, zhao2023survey, giancola2018soccernet,deliege2021soccernet,yuan2021spatio} has advanced with natural language processing (NLP) and computer vision (CV), multimodal large language models (MLLMs)~\cite{achiam2023gpt, hurst2024gpt} have made question-answering (QA) tasks increasingly valuable for sports comprehension.
Sports-focused QA \cite{liu2020liveqa, srivastava2022beyond, xia2024sportqa, li2024sports, xia2024sportu} are typically divided into two formats: text-based and video-based. The former focuses on numerical data, rules, and context, while the latter emphasizes action/movement understanding and reasoning. However, these datasets largely prioritize tactics and strategy, which contrasts with the needs of artistic sports like figure skating, where assessing action quality and overall performance is crucial.
 
On the other hand, existing MLLMs that handle long videos struggle to focus on local details~\cite{li2025llama, fan2025videoagent, zhang2024flash}, while MLLMs designed for short videos have difficulty processing long videos~\cite{zhang2023video, lin2023video, jin2024chat}.
To overcome these limitations and enhance traditional skeleton-based figure skating datasets, we introduce the use of motion data. Unlike skeleton data, motion data can more accurately capture intricate and subtle movements, such as spins, jumps, and fluid transitions, that are essential for evaluating the quality of performance. 
It also captures the temporal relationships and continuity between movements, making it more suitable for action recognition and classification tasks that rely on understanding movement fluidity and overall coordination. 
Motion data typically includes information about the body's shape and appearance, such as joint movements, muscle extensions, and overall posture contours. This data provides a more realistic representation of the athlete's movements and poses, preserving the visual aesthetics of a figure skating performance. 
Our contributions are summarized as follows:

\begin{itemize}
    \item Our work introduces FSAnno, the first fine-grained annotated dataset for figure skating, supporting multi-role, multi-modal, and multi-level tasks. Unlike previous datasets designed for single tasks, FSAnno provides comprehensive annotations across text, visual, and audio modalities, along with various evaluation schemes, making it particularly suitable for training general-purpose models, especially LLMs. 
    \item We developed an evaluation benchmark, FSBench, which provides a rich assessment platform for potential methods, particularly LLMs. FSBench-Text includes multiple-choice questions with human-annotated explanations, aimed at understanding rule-specific and event-specific information. FSBench-Motion contains multi-modal action and QA pairs, supporting tasks from analyzing individual technical actions to generating comprehensive performance commentary. This structured task design allows FSBench to evaluate a model's human-like understanding more effectively. To the best of our knowledge, this is the first dataset that emphasizes the artistic and technical complexity of figure skating, offering a holistic view of the sport.
    \item Existing methods perform inadequately on FSBench, highlighting the limited understanding of artistic sports like figure skating in current LLMs. For certain tasks, we designed instruction-tuning datasets based on FSAnno and proposed the SkateLLM. Test results show that our fine-grained, multi-modal annotations significantly enhance the LLMs' capabilities.
\end{itemize}

\begin{figure*}[t]
\begin{center}
\includegraphics[width=14cm]{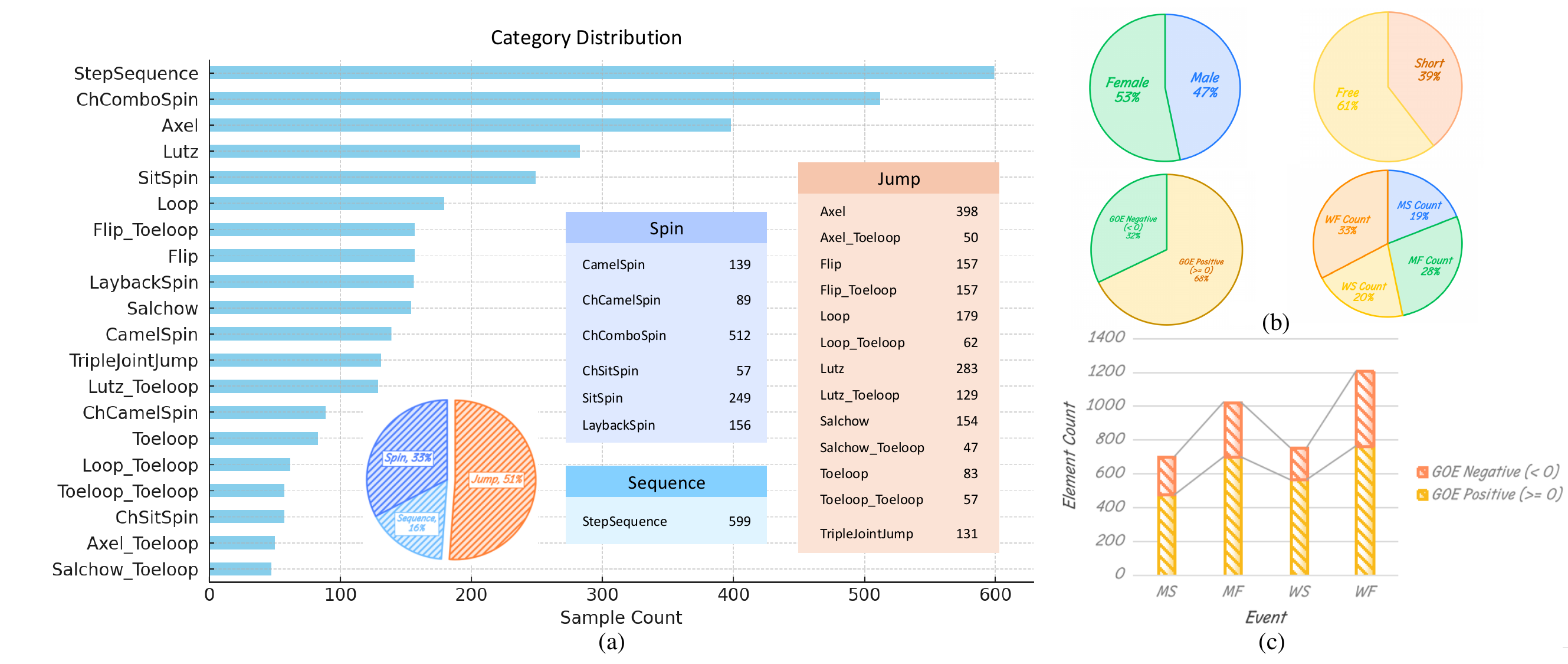}
\end{center}
\vspace{-2em}
\caption{Data Statistics. (a) The figure skating elements are divided into three main categories: Jump, Spin, and Sequence, which are further divided into 20 subcategories. The bar chart shows the category distribution. (b) The pie chart shows the distribution of FSBench elements by gender; Men Short Program (MS), Men Free Skating (MF), Women Short Program (WS), Women Free Skating (WF); short program and free skate; and the positive and negative values of the $GOE$ scores. (c) The stacked bar chart shows the distribution of $GOE$ scores across different program types.}
\vspace{-1em}
\label{fig:data}
\end{figure*}
\section{Related Work}
\label{sec:related_work} 

\textbf{Figure Skating.}
Figure skating, with its complex technical maneuvers, has attracted numerous researchers. Many datasets have been proposed to address related challenges, which can be categorized into three areas: 
1) Action Recognition. The FSD-10 dataset~\cite{liu2020fsd} extends coverage to 10 different movements across male/female events. 
2) Temporal Action Segmentation. MCFS~\cite{liu2021temporal} shifts focus toward temporal action segmentation, featuring both RGB and skeletal data, alongside granular action category and temporal annotations. 
3) Action Quality Assessment/Scoring.
MIT-Skating~\cite{pirsiavash2014assessing} comprises 150 videos annotated with corresponding scores. 
The FisV dataset~\cite{xu2019learning} has amassed 500 videos of women's singles figure skating, each annotated with its Total Element Score (TES) and Total Program Component Score (PCS). 
Expanding significantly, the FS1000 dataset~\cite{xia2023skating} encompasses over 1,000 videos across eight types of programs, incorporating audio modality (music) and seven distinct scoring metrics. 
Most recently, the OlympicFS dataset~\cite{du2023learning} not only includes annotations for figure skating scoring but also professional commentaries on the movements to facilitate the learning of semantic representations. However, these datasets are primarily designed for single tasks and lack comprehensive annotations, which limits their applicability in multi-task scenarios. Few studies and datasets have paid attention to the characteristics of ``artistry'' in figure skating. Furthermore, most datasets are sourced from elite-level figure skating competitions, resulting in a limited number of negative samples. Nevertheless, negative samples are essential for a range of tasks, including description, commentary, scoring, and training guidance. To support further advancements in this field, there is a pressing need for a fine-grained, multi-task dataset.

\begin{figure*}[t]
\begin{center}
\includegraphics[width=13cm]{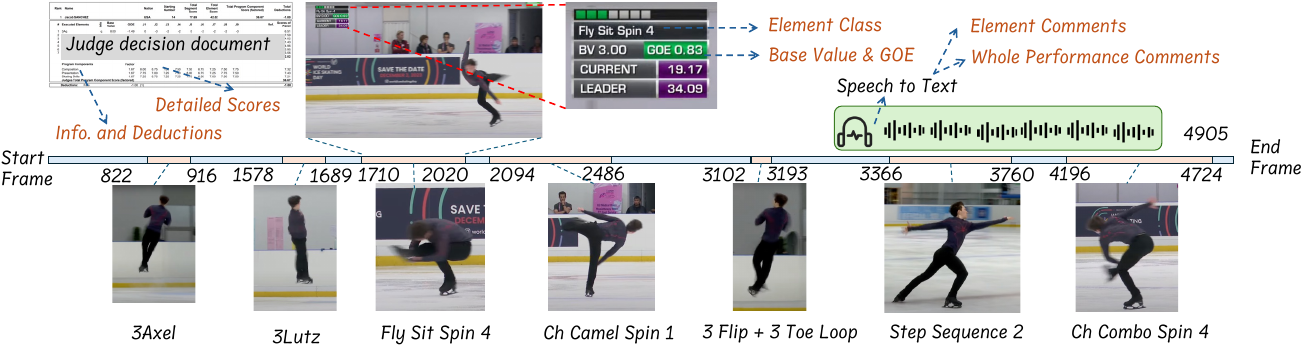}
\end{center}
\vspace{-1.5em}
\caption{Our annotations are derived from three sources: official judging reports, visual information, and audio commentary. These three official sources provide objective, multi-dimensional, fine-grained, and multi-task-oriented annotations for FSAnno and FSBench.}
\vspace{-1em}
\label{fig:anno}
\end{figure*}

\textbf{Sports Understanding.}
In recent years, with the integration of NLP and CV, QA has become a mainstream benchmark format~\cite{li2024mvbench, zhu2023mme, li2024seed}, and several benchmarks have been developed to evaluate tasks in sports~\cite{liu2020liveqa, srivastava2022beyond, xia2024sportqa, li2024sports, xia2024sportu}. 
SportQA~\cite{xia2024sportqa} provides a text-only dataset with over 70,000 multiple-choice questions designed across three levels: basic sports knowledge, rules and tactics comprehension, and complex scenario-based reasoning. 
SPORTU~\cite{xia2024sportu} expands beyond text by introducing a multimodal benchmark that incorporates both text-based and video-based tasks. SPORTU mainly centers on ball sports and evaluates understanding at complexity, such as rule comprehension, strategy, and scenario-based reasoning.
Sports-QA~\cite{li2024sports} includes video data from various sports and generates approximately 94,000 QA pairs, covering descriptive, temporal, causal, and counterfactual questions.
However, SportQA~\cite{xia2024sportqa} and SPORTU~\cite{xia2024sportu} focus exclusively on ball sports, and while Sports-QA~\cite{li2024sports} includes some gymnastics content, which primarily frames questions from a general action-understanding video perspective, focusing on temporal and causal aspects. 
These datasets lack focus on technical skill, emotional expression and the fluidity of performance.
This creates an opportunity for developing benchmarks that challenge models to capture the intricate mechanics of athlete movements and interpret more sophisticated aspects of sports action beyond ball sports.

\textbf{Motion MLLMs.}
Recent advancements in MLLMs have led to numerous approaches focused on motion, typically categorized into motion generation~\cite{jiang2023motiongpt, wu2024motionagent} and motion understanding~\cite{jiang2023motiongpt, zhou2024avatargpt, chen2024motionllm}. For instance, recent frameworks such as MotionGPT~\cite{jiang2023motiongpt}, AvatarGPT~\cite{zhou2024avatargpt}, and MotionLLM~\cite{chen2024motionllm} LLMs to understand human poses and motions. However, these approaches primarily target general human motion captioning and lack the specialized spatial-temporal awareness and reasoning abilities required for detailed sports-level understanding.
\section{FSBench}

FSAnno aims to advance artistic sports understanding such as figure skating, making it more aligned with real-world applications. In this section, we provide a detailed introduction to FSAnno and FSBench, including the key challenges of the dataset, its tasks and scalability, annotation methods, and category definitions. We also present the dataset properties and statistics of the FSAnno.

\subsection{Key Challenges}

We aim to begin by building and sharing a new figure skating database, hoping to attract more community attention to the automation of figure skating commentary and scoring. Several unprecedented challenges need to be addressed to construct FSAnno, as it differs from previous datasets. Not only does it target tasks closer to real-world scenarios, but we also hope it will bridge multimodal data, providing a richer foundation for subsequent research based on LLMs. 1) \textbf{\textit{How to help better understanding of the artistic sport of figure skating?}} Taking the scoring process of real-world judges as an example, the panel scores each technical move performed by the athlete, rates the overall performance to derive the program component score, and deducts points for mistakes such as falls, incorrect steps, or irregular movements. The combination of these three factors constitutes the final score. Previous methods often passed the input directly through a black-box model to obtain the final score, leading to a lack of interpretability that is insufficient to support automated scoring methods. We have decomposed the task into two levels: individual actions and entire performance. For the individual actions level, we have added annotations for assessment, commentary, and anomalies, allowing the dataset to contain sufficient information to meet our objectives. 2) \textbf{\textit{How to annotate the data?}} For non-professionals, manually identifying and segmenting figure skating moves is extremely challenging, especially since FSAnno requires fine-grained annotations at two levels: individual actions and entire performances. Our annotations for single action recognition, long video segmentation, and overall performance scoring align with the granularity and standards of previous datasets~\cite{liu2021temporal}. For additional annotations required by other tasks, we use the official technical documents of figure skating as a guide to ensure the accuracy of the marked data (details in the Data Annotation section). 3) \textbf{\textit{How to obtain artistic annotations?}} Objective scores are derived from the judges' reports, including ``Performance'' and ``Composition''. In addition to evaluations of artistry extracted from audio commentary, we synthesized additional artistic comments by integrating ``Performance'', ``Composition'', and ``$GOE$'' scores.
4) \textbf{\textit{How to protect athletes' privacy and avoid information leakage?}} MLLMs have vast knowledge reservoirs. As an evaluation benchmark, FSBench needs to prevent MLLMs from using prior knowledge to score and comment on performances by directly identifying competitions through athletes' appearances or additional information in video frames. Since different MLLMs have significantly varying prior knowledge, we have extracted identity-free data from motion and skeleton modalities to use cleaner data as the basis for a fairer evaluation while protecting athletes' privacy.

\begin{table*}[htp]
  \centering
  \vspace{-1em}
  \caption{Task examples of FSBench.}
  \vspace{-1em}
  \footnotesize
  \begin{tabular}{l|c|l}
    \Xhline{1pt}
    & \textbf{Tasks} &  \textbf{Examples} \\
    \hline
    \multirow{2}{*}{\makecell[l]{\textbf{Prior knowledge}}} & Rules & \makecell[l]{\textcolor{lightblue}{What distinguishes a `combination' jump from a `sequence' jump?} \\
    (A) Combination jumps involve turns between jumps\\
    (B) Sequence jumps are always of the same type\\
    (C) Combination jumps have no steps between the jumps\\
    (D) Sequence jumps require a change of edge}  \\
    \cline{2-3}
    & Event Information &  \makecell[l]{\textcolor{lightblue}{In which city was the World Figure Skating Championships 2013 held?} \\
    (A) Sarajevo (B) Gangneung (C) Birmingham (D) London}\\
    \Xhline{1pt}
    \multirow{3}{*}{\makecell[l]{\textbf{Individual actions}}} & Action Recognition / Caption &  \makecell[l]{\textcolor{lightblue}{Which of the following three categories does the figure skating move}\\
    \textcolor{lightblue}{in the [input motion] belong to?} \\
    (A) Spin (B) Sequence (C) Jump\\
    \textcolor{lightblue}{To which specific category does it belong?}\\
    (A) ChoreoSequence1 (B) ChComboSpin4 (C)FlyCamel-Spin4 \\
     (D) StepSequence3 (E) 2Axel (F) 3Loop (G) 3Flip (H) 3Axel\\
    (I) 3Lutz (J) 3Lutz\_3ToeLoop  \\
    \textcolor{lightblue}{Can you translate this [input motion] to text?} }\\
    \cline{2-3}
    & Single Action Assessment &  \makecell[l]{\textcolor{lightblue}{Please assess whether the GOE value is positive or negative} \\
    \textcolor{lightblue}{for the action in [input motion].} \\
    (A) Positive (B) Negative (C) 0\\
    \textcolor{lightblue}{Can you provide the specific GOE score for [input motion]?}} \\
    \cline{2-3}
    & Single Action Commentary Generation &  \makecell[l]{\textcolor{lightblue}{As a professional figure skating commentator, please comment} \\
    \textcolor{lightblue}{on the figure skating moves in [input motion]}.}\\
    \Xhline{1pt}
    \multirow{3}{*}{\makecell[l]{\textbf{Entire performances}}} & Action Segmentation &  \makecell[l]{\textcolor{lightblue}{What figure skating moves are included in [input motion]?} \\
    \textcolor{lightblue}{Please list them in order.}}\\
    \cline{2-3}
    & Performance Assessment &  \makecell[l]{\textcolor{lightblue}{As a professional figure skating judge, please score each technical} \\
    \textcolor{lightblue}{execution and provide scores for each dimension of the complete}\\
    \textcolor{lightblue}{performance, along with the total score.}}  \\
    \cline{2-3}
    & Performance Commentary Generation &  \makecell[l]{\textcolor{lightblue}{How would you evaluate this [figure skating] performance?} \\
    \textcolor{lightblue}{What are your thoughts on this [figure skating] presentation?}\\
    \textcolor{lightblue}{Could you share your opinions on this [figure skating] performance?}}  \\
    \Xhline{1pt}
  \end{tabular}
  \label{tb:task_example}
\end{table*}

\subsection{Task Definition and Scalability}

As described in the introduction, most existing datasets and methods focus on single tasks, and although the modalities of the datasets are not uniform, we find that in most cases the annotations are not affected. To facilitate community expansion and unified use of these datasets, we have aligned the existing tasks and annotations with those of previous datasets while expanding the scale of FSBench. As shown in Table~\ref{tb:task_example}, FSBench ultimately includes six main tasks across two levels and a prior knowledge test, as outlined below.

\textbf{Prior knowledge test}: We collect more than 4,200 multiple-choice questions on rules and event information, which can be used to test the LLMs' prior knowledge of figure skating.

\textbf{Individual actions}: 1) Action Recognition: Identifying and classifying individual basic figure skating moves. 2) Single Action Assessment: Assessing the quality of individual figure skating moves. 3) Single Action Commentary Generation: Generating text-based evaluations and corresponding comments for a given action.

\textbf{Entire performances}: 1) Action Segmentation: A complete performance includes a combination of multiple actions; the goal of this task is to segment these sequences of connected technical actions. 2) Performance Assessment: Scoring a complete performance from multiple dimensions. 3) Performance Commentary Generation: Generating text-based evaluations and corresponding comments for a complete performance.

In addition to the tasks mentioned above, we highlight that the fine-grained annotations in FSBench are sufficient to support its extension to other tasks, such as anomaly action recognition (e.g., falls) and guidance action generation (human motion generation). FSBench offers rich scalability and provides a broad platform for the community's research on figure skating.

\begin{table*}[htp]
    \centering
    \caption{The attributes comparison of FSBench with other widely used datasets for figure skating analysis and understanding. (Class: Action Class, AA: Action Assessment, AC: Action Comments, AS: Action Segmentation, PS: Performance Score, PC: Performance Comments, AR: Action Recognition, LS: Long-Video Scoring, A: Audio, V: Video, M: Motion, S: Skeleton, T: Text.) The number of videos corresponds to either the number of elements or performances.}
    \begin{tabular}{|c|c|c|c|c|c|c|c|c|c|c|}
        \hline
        \multirow{2}{*}{Datasets} & \multirow{2}{*}{Task} & \multirow{2}{*}{\#Video} & \multirow{2}{*}{Length} & \multirow{2}{*}{Feature} & \multicolumn{6}{c|}{Annotations} \\
        \cline{6-11}
         & & & & & Class & AA & AC & AS & \#PS & PC \\
        \hline
        FSD-10 & AR & 1484 & 10h & V & \checkmark & \texttimes & \texttimes & \texttimes & \texttimes & \texttimes \\
        MCFS & AS & 271 & 5h & V & \checkmark & \texttimes & \texttimes &  \checkmark & \texttimes & \texttimes  \\
        MIT-Skate & LS & 171 & 8h & V & \texttimes & \texttimes & \texttimes & \texttimes & 1 & \texttimes \\
        FisV & LS & 500 & 24h & V & \texttimes & \texttimes & \texttimes & \texttimes & 2 & \texttimes \\
        FS1000 & LS & 1604 & 91h & A+V & \texttimes & \texttimes & \checkmark & \texttimes & 7 & \texttimes \\
        OlympicFS & LS & 200 & - & V+T & \texttimes & \texttimes & \checkmark & \texttimes & 7 & \texttimes \\
        FSBench & AR+AS+LS+AC+PC & 783 & 76h & A+V+M+S+T & \checkmark & \checkmark & \checkmark & \checkmark & 7 & \checkmark \\
        \hline
    \end{tabular}
    \vspace{-1em}
    \label{tab:dataset}
\end{table*}

\subsection{Dataset Construction}

\textbf{Data Preparation}.
Unlike previous datasets~\cite{liu2021temporal, xia2023skating, du2023learning} that only sourced data from the best world-class events, we have collected data from 11 competitions at the ISU Grand Prix of Figure Skating and the ISU Junior Grand Prix of Figure Skating, encompassing over 783 complete figure skating performances. The programs include four categories: Men's Short Program, Women's Short Program, Men's Free Skating, and Women's Free Skating. This diversity allows our dataset to include samples of athletes across different age groups and levels of experience. Additionally, we simultaneously collected authoritative judges' reports corresponding to each event.

The collected videos typically contain unedited, complete records of the competitions ranging from 1 to 6 hours, including warm-ups, complete figure skating performances, highlight replays, and score waiting periods. Unlike previous datasets that only required footage of the performances, FSAnno also gathers content from highlight replays and score waiting periods to capture commentary from human commentators and evaluations of specific actions.

For edited figure skating performance videos, we utilize 4DHumans~\cite{goel2023humans} to extract corresponding motion data and HRNet~\cite{sun2019deep,xiao2018simple} to estimate and extract skeleton data. In addition to these two modalities, we also provide links to the original RGB videos to facilitate research on figure skating in RGB format by the research community.

\textbf{Data Annotation}. As shown in the figure~\ref{fig:anno}, the dataset annotations include not only score annotations from judging reports but also element segmentation annotations assisted by video cue boxes, as well as commentary annotations extracted from audio. As mentioned earlier, our annotations in FSBench are guided by official authoritative judges' reports, and for existing tasks, we have aligned our annotations as closely as possible with the standards of previous datasets. For each individual action, we have matched and annotated the category and $GOE$ (Grade of Execution) scores for single actions based on the referee's report. Each technical element has a base value, and the $GOE$ adjusts this base score up or down depending on the quality of the athlete's execution of the element. The $GOE$ values range from $-5$ to $+5$, with higher scores indicating better execution quality and lower scores indicating flaws in execution. The commentary comes from the announcers' evaluations during the competition and highlights replays. After collecting the audio information, we used Whisper~\cite{radford2023robust} to convert the speech to text. We then assigned the commentators' remarks to the corresponding actions based on timestamps. It's important to note that not every action has matching commentary; this is because commentators generally only remark on actions that are either executed exceptionally well or involve noticeable errors. 

For each entire performance, we annotated the start and end frames of actions for the segmentation task, and multi-dimensional scores for the assessment task, and provided a summarizing commentary for each performance. Based on the judges' reports, our scoring annotations include Technical Element Scores (TES) and Program Component Scores (PCS). TES is the sum of all technical elements' base scores adjusted by their $GOE$. PCS evaluates five dimensions: Skating Skills (SS), Transitions (TR), Performance (PE), Composition (CO), and Interpretation of the Music (IN). The final score is determined by both TES and PCS. Unlike the annotation scheme for commentary on individual actions, where commentators usually evaluate specific actions and often do not provide a summarizing assessment of the entire performance, we used the Large Language Model to process the collected commentary from commentators. This approach allows us to derive more comprehensive and summarizing comments on the entire performance.

\subsection{Dataset Statistics and Properties}

The dataset provides not only the figure skating performance videos but also highlights replay videos and audio from the results waiting period, which includes commentary by announcers. The videos are processed to be identity-free and are available in two different modalities: motion and skeleton. The average duration of the short program videos is 2 minutes and 52 seconds, while the free skating videos average 4 minutes and 13 seconds in length. Information such as the distribution of classes in the dataset, percentages of different dimensions, and other relevant details are illustrated in Figure~\ref{fig:data}. A comparison of these statistics with other datasets can be found in Table~\ref{tab:dataset}.

\subsection{Evaluation Method}

For the multiple-choice questions in the figure skating understanding task, we use accuracy as the evaluation metric to directly measure the model's ability to predict the correct options. However, for open-ended question-answering tasks, we use AutoDQ~\cite{wang2024tarsier} as the evaluation metric. Compared to traditional evaluation metrics (such as BLEU\cite{papineni2002bleu}, GLEU\cite{mutton2007gleu}, and METEOR\cite{banerjee2005meteor}), AutoDQ is more suitable for assessing complex, detailed descriptions and technical analyses of figure skating movements. This is because traditional metrics generally focus on word matching or surface structure similarity, which often fails to capture the semantic nuances, contextual accuracy, and technical details of movements, posture changes, and technical components.

The AutoDQ evaluation process consists of two main stages: In the first stage, key events are extracted from the ground-truth description and the LLM-generated description. These events represent core elements of figure skating movements, such as the edge of take-off, rotation direction, aerial posture, and landing details. Each event reflects essential technical information or changes in posture within the descriptions, forming the basis for comparing the model-generated and human descriptions. In the second stage, AutoDQ calculates recall and precision based on the matching of events and measures the alignment between the model-generated description and the ground-truth.
\section{SkateLLM}

The performance of MLLMs is significantly influenced by instruction-tuning data, especially when previous MLLM evaluations have involved minimal training on figure skating-related datasets, resulting in a limited understanding of this domain. With this in mind, we conduct instruction tuning on the fundamental task of motion captioning, building on the Motion-GPT model.
\subsection{Instruction-Tuning Data}
We designed an instruction-tuning dataset that incorporates previous datasets to enrich the data. This approach not only preserves the model’s general motion understanding capabilities but also enhances its comprehension of figure skating-specific actions. General motion understanding data is sourced from established datasets~\cite{guo2022generating,plappert2016kit}. Figure skating motion description data is derived from FSAnno, with key caption ground truth annotations informed by referee reports. We crafted templates manually which follow a structured approach that ensures each template accurately conveys the technical aspects of figure skating movements while reflecting the diversity of artistic expression. We use GPT4 to generate data in batches according to the template. 

Specifically, we used GPT-4 to generate descriptions for different element categories, creating templates that included technical details like take-off edge, rotation, air position, and landing control. These templates were enhanced with artistic evaluations based on $GOE$ scores: high scores received positive artistic commentary on fluidity and expressiveness, while lower scores included constructive critiques. This approach helps the model better understand technical and artistic aspects and perceive performance quality variations.

\subsection{Coordinated Multi-modal Training} 
FSAnno provides scalable motion and skeleton data, facilitating instruction-tuning dataset expansion. To bridge the visual-linguistic gap, we used a two-phase training strategy: aligning visual embeddings with LLM text embeddings and optimizing with visual instructions. Cross-entropy loss was applied for token prediction, with the cross-modal projector pre-trained for one epoch.
According to~\cite{he2024efficient}, the LoRA~\cite{hu2021lora} technique can effectively mitigate catastrophic forgetting in smaller models. So during the instruction fine-tuning phase, we conducted one epoch of training on the cross-modal projector and the main body of the LLM.
\section{Experiment}

\subsection{LLM prior knowledge evaluation}

FSBench includes a task specifically for figure skating knowledge, designed to assess the depth of understanding of foundational LLMs in this field. It is further divided into two parts. The first part consists of 500 specialized questions on figure skating rules, covering knowledge such as scoring criteria and technical details. The second part includes 700 questions randomly selected from a pool of 3,500, focusing on objective information about events, such as competition locations, years, and champions. The purpose of designing this task is to evaluate the knowledge base and response accuracy of LLMs in the specialized domain of figure skating. The testing uses the ``5 shot strategy''. 

\begin{table}[htp]
\centering
\caption{Comparison of LLMs' accuracy (\%) on the FSBench prior knowledge test and other institutions' tests.}
\small
\begin{tabular}{|l|c|c|c|}
\hline
Method   & Event Info.  & Rules & Quiz \\    
\hline
    \hline
GPT3.5-turbo & 59.0   & 64.8  & 72.7\\
GPT4 & 73.0 & 78.0  &  87.9 \\
LLaVA 13B & 47.0  & 51.4  & 63.6 \\
    \hline
\end{tabular}
\label{tb:FSBench_prior}
\end{table}

We also collected figure skating knowledge quiz data from publicly accessible data sources that classifies respondents into five levels based on their scores: Armchair Athlete, Rookie Runner, Midfield Maestro, All-Star Striker, and Legendary Champion. According to this classification, only GPT-4 reached the ``All-Star Striker'' level, while other large language models performed relatively average. This result aligns with the prior knowledge test results on FSBench as shown in Table~\ref{tb:FSBench_prior}, indicating that current large language models still lack sufficient understanding in arts-related, highly specialized sports like figure skating. Due to page limits, please refer to the supplementary materials for more details.

\subsection{Figure Skating Element Description}
The construction of FSBench provides fertile ground for training and testing with a comprehensive, fine-grained, and more professional approach. However, existing video-based and motion-based LLMs are generally unable to comprehensively address FSBench. In our paper, we primarily focus on motion-based methods, most of which lack the capability for a complete evaluation on FSBench. Nevertheless, we have completed partial tests as much as possible. For details on the evaluation results of existing video-based and motion-based LLMs on FSBench, please refer to the supplementary materials and our official repository.

\begin{figure}[t]
\begin{center}
\includegraphics[width=7.5cm]{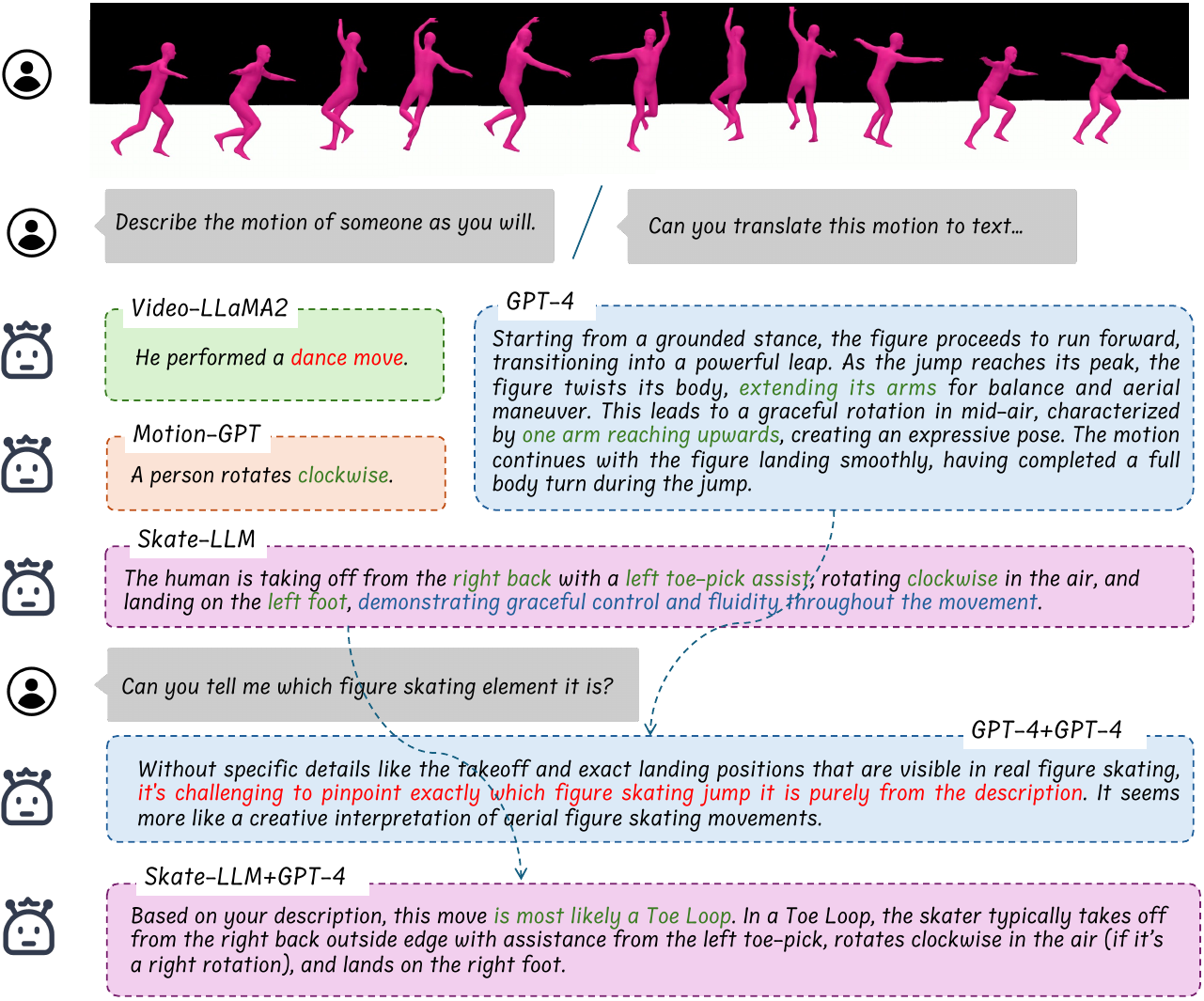}
\end{center}
\vspace{-1em}
\caption{Compared to other large language models, SkateLLM's descriptions of figure skating elements are more focused on the technical movements themselves and artistic evaluations. Additionally, based on the more professional captions it generates, other LLMs (such as GPT-4) have a higher probability of correctly inferring the category of the element.}
\vspace{-1.5em}
\label{fig:SkateLLM_case}
\end{figure}

In our paper, we focus temporarily on the most fundamental task: generating more suitable descriptions for figure skating elements. This foundation allows us to further delve into more complex tasks like action classification and commentary. As shown in Figure~\ref{fig:SkateLLM_case}, many existing methods, including GPT-4, are actually insufficient in generating concise, professional descriptions specifically tailored to figure skating movements. Leveraging the rich, detailed annotations and data in FSBench, we developed SkateLLM through instruction fine-tuning on MotionGPT~\cite{jiang2023motiongpt}. As illustrated in Figure~\ref{fig:SkateLLM_case}, compared to other methods, the captions generated by SkateLLM are more focused on information related to figure skating movements, particularly descriptions of key body parts and directional details.

\begin{figure}[t]
\begin{center}
\includegraphics[width=7.5cm]{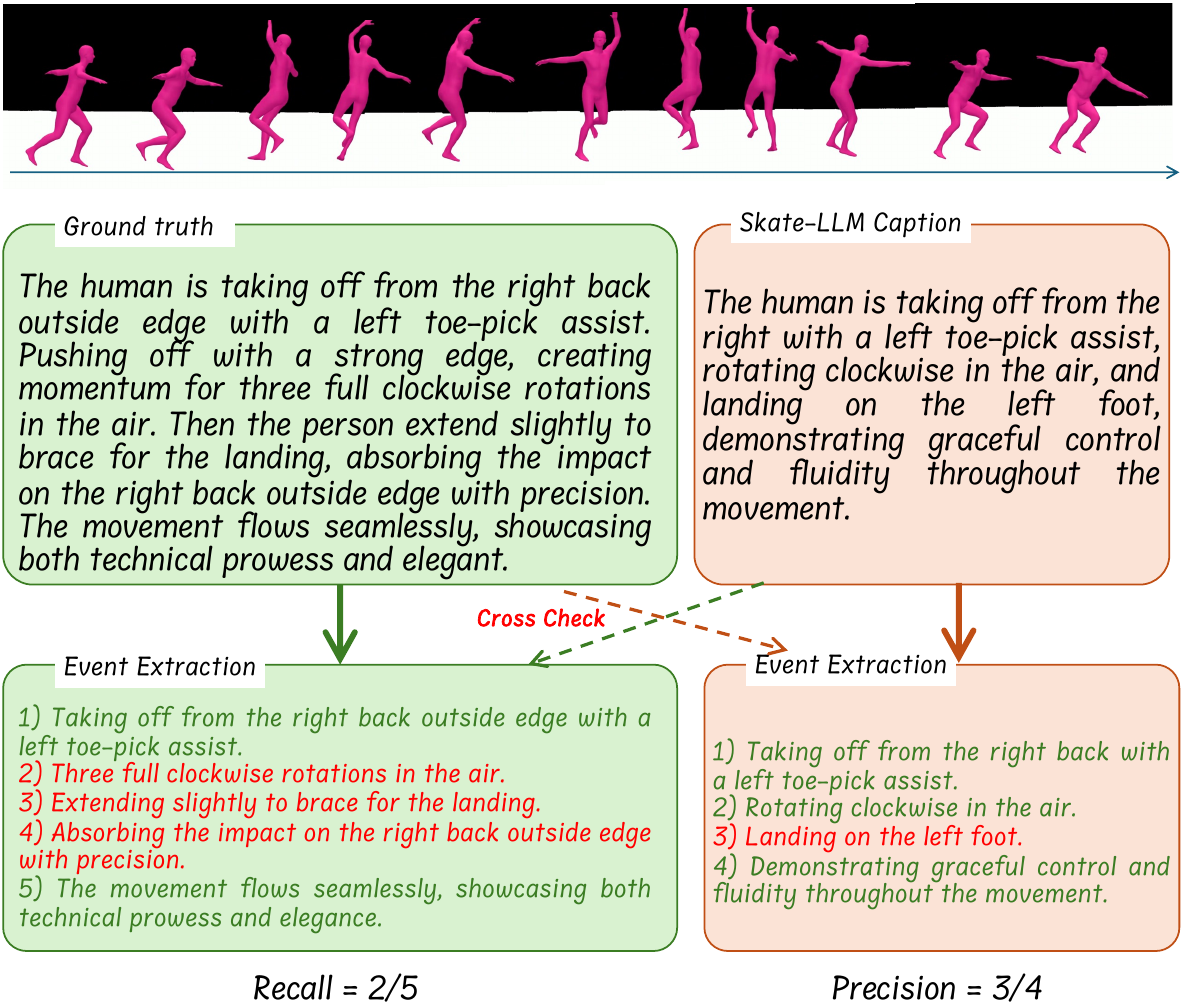}
\end{center}
\vspace{-1em}
\caption{Evaluation pipeline using AutoDQ. For event extraction and cross-checking, we use GPT-3.5-turbo. These results can support more fine-grained evaluation.}
\vspace{-1em}
\label{fig:recall_precision}
\end{figure}

According to the evaluation metrics for text generation tasks in FSBench, we use AutoDQ to evaluate SkateLLM. Figure~\ref{fig:recall_precision} provides an example of the evaluation process. Key events are extracted separately from both the ground truth and the captions generated by SkateLLM. By comparing the key events extracted from the captions generated by SkateLLM with the ground truth, we obtain the precision value. Similarly, by comparing the captions generated by SkateLLM with the key events extracted from the ground truth, we calculate the recall rate. This method allows us to more accurately assess whether SkateLLM has thoroughly understood the critical body parts and movement forms involved in figure skating elements.

\begin{table}[htp]
\centering
\caption{Comparison of Motion-based LLMs via AutoDQ-based metrics.}
\vspace{-1em}
\small
\begin{tabular}{|l|c|c|c|}
\hline
Method   & F1($\uparrow$)  & Recall($\uparrow$) & Precision($\uparrow$) \\    
\hline
    \hline
Motion-GPT & 7.1   & 3.772  & 27.456\\
SkateLLM & 38.0 & 58.333  &  61.622 \\
    \hline
\end{tabular}
\vspace{-1.5em}
\label{tb:AutoCQ}
\end{table}

Table~\ref{tb:AutoCQ} shows the results of different motion-based LLMs evaluated using AutoDQ. It can be observed that the performance of Motion-GPT, which has not undergone instruction fine-tuning, is unsatisfactory despite being trained on various motion datasets. SkateLLM exhibits higher precision but lower recall, which we believe is due to the significant similarity between many figure skating elements, especially among movements that differ only by the number of rotations. Although this information is included in the ground truth, SkateLLM struggles with this aspect due to the rapid rotation speed and the common challenge of repetitive counting in motion domains. The high precision suggests that SkateLLM performs well in recognizing broad categories, but a deeper understanding is still required for finer distinctions. In the supplementary materials, we present additional evaluations on FSBench as well as more example demonstrations.
\section{Conclusion}

We introduce FSAnno, a fine-grained, multi-modal dataset for figure skating, designed to capture both technical and artistic elements. We construct FSBench to evaluate model performance, exploring limitations in current LLMs' understanding of artistic sports. Instruction-tuning with FSAnno on specific models for some tasks show significant improvements, highlighting FSAnno and FSBench’s potential to enhance LLM comprehension of figure skating. Future work will focus on training a MLLM fully optimized for FSAnno, enabling it to effectively address the diverse tasks in figure skating understanding.

\clearpage
\bibliographystyle{ieeenat_fullname}
\bibliography{main}

\end{document}